# Knowledge Base of an Expert System Used for Dyslalic Children Therapy

Ovidiu-Andrei SCHIPOR, Ştefan-Gheorghe PENTIUC, Doina-Maria SCHIPOR
"Stefan cel Mare" University of Suceava
str.Universitatii nr.13, RO-720229 Suceava
schipor@eed.usv.ro, pentiuc@eed.usv.ro, vmdoina@yahoo.com

*Abstract* — In order to improve children speech therapy, we develop a Fuzzy Expert System based on a speech therapy guide. This guide, write in natural language, was formalized using fuzzy logic paradigm. In this manner we obtain a knowledge base with over 150 rules and 19 linguistic variables. All these researches, including expert system validation, are part of TERAPERS project (financed by the National Agency for Scientific Research, Romania).

## I. INTRODUCTION

The main objectives of speech therapy expert system develop by our team are [1]:
- personalized therapy (the therapy must be in according with child's problems level, context and possibilities);
- speech therapist assistant (the expert system offer some suggestion regarding what exercises are better for a specific moment and from a specific child);
- (self) teaching (when system's conclusion is different that speech therapist's conclusion the last one must have the knowledge base change possibility).

## II. EXPERT SYSTEM ROLE IN ASSISTED THERAPY

A full computer based speech therapy system must contain at least following items [2]:
- monitor program (installed on speech therapist computer, helps on children data management);
- expert system (based on children related information's obtain a personalized therapy path);
- exercises set (for cabinet and home use);
- home training possibility (exercises on personal computer or mobile device).

Architecture of our developed speech therapy system is presented in figure 1.

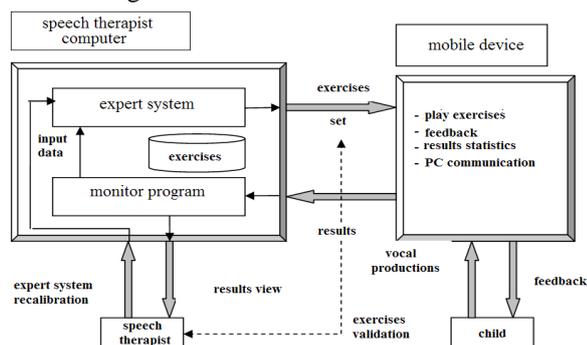

Figure 1 – Implemented speech therapy system architecture

Mobile device receives exercises (created on speech therapist computer). The role of this device is to create a virtual-operative interface between teacher and child (speech therapy work at home):
- play exercises in a game manner;
- offer real time feedback for child;
- realize statistics related at current state of the subject;
- implement specific communication protocols.

Expert system receives information from monitor program (children personal data and pronunciation scores) and from speech therapist (recalibration of knowledge database). It offers specific information referring to optimum next therapeutically steps (number of training sessions, number of homework, collaboration with specialized doctor, etc).

In figure 2 are presented therapeutically steps and necessary distributed knowledge bases [3].

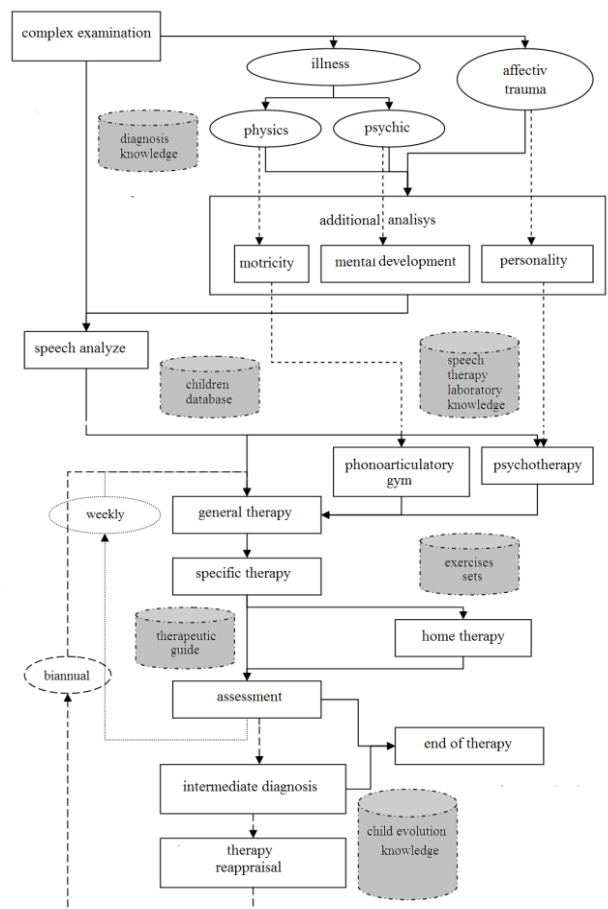

Figure 2 – Therapy necessary knowledge database

In according with [4], speech therapy software can help speech problems diagnostic, can offer a real-time, audio-visual feedback, can improve analyze of child progress and can extend speech therapy at child home. Specific therapy information can be found in:
- dyslalia therapeutically guides;
- speech therapy centers experience;
- dyslalia exercises sets;
- historical therapy data.

### III. EXPERT SYSTEM – FUZZY APPROACH

With fuzzy approach we can create a better model for speech therapist decisions. Fuzzy logic has ability to create accurate models of reality. It's not an "imprecise logic". It's a logic that can manipulate imprecise aspects of reality. In the latest years, many fuzzy expert systems were developed [5], [6], [7].

In the next set of pictures, we present an example of fuzzy inference. There are three input linguistic variables (speech problems level – figure 3, family implication – figure 4 and children age – figure 5) and one output linguistic variable (weekly session number – figure 6). We consider five fuzzy rules and, base on these rules, we illustrate specific fuzzy result (figure 7). If system user wants a crisp value, deffuzification is a good solution (figure 8).

To express a number in words, we need a way to translate input numbers into confidences in a fuzzy set of word descriptors, the process of *fuzzification*. In fuzzy math, that is done by *membership functions* [8].

Defuzzification is the reverse process of fuzzification. We have confidences in a fuzzy set of word descriptors, and we wish to convert these into a real number.

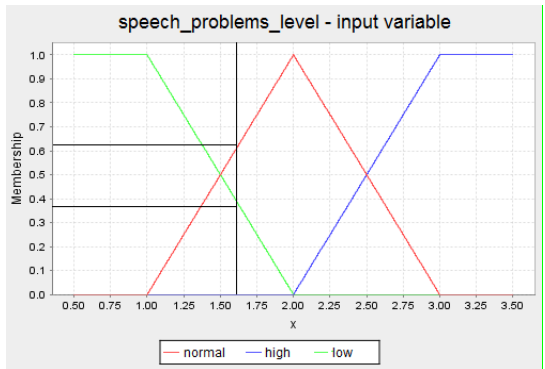
Figure 3 – Speech_problems_level language variable

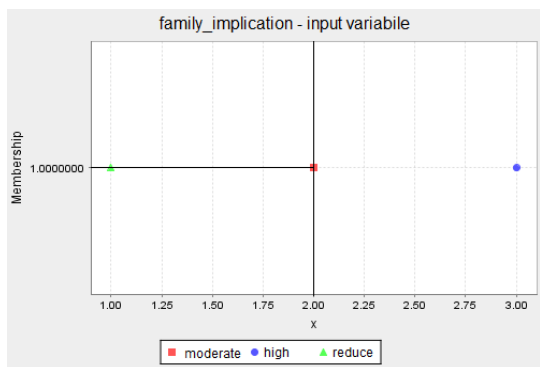
Figure 4 – Family_implication language variable

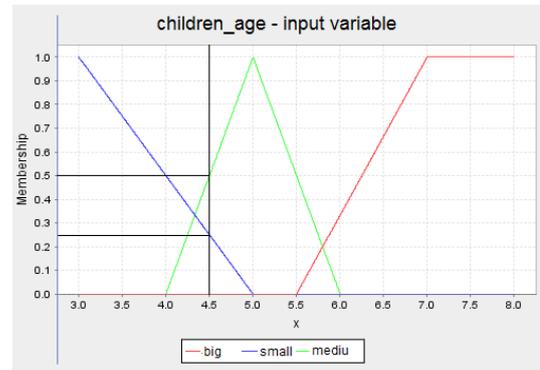
Figure 5 – Children_age language variable

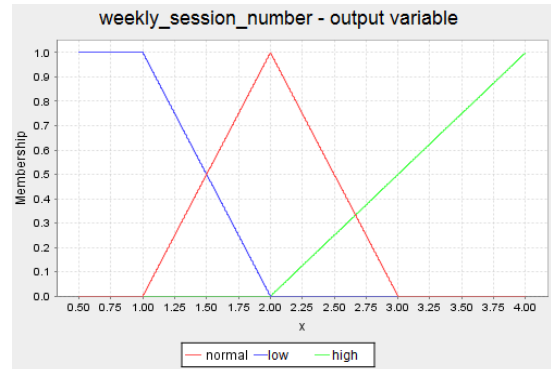
Figure 6 – Weekly_session_number language variable

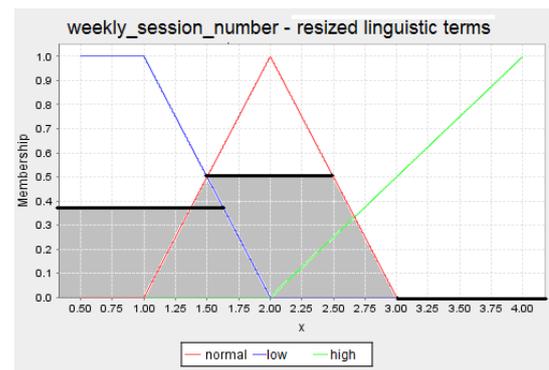
Figure 7 – Obtain a fuzzy result

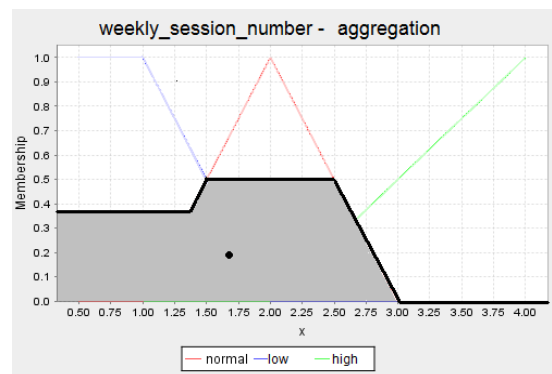
Figure 8 – Obtain a crisp value

First three variables have following representation:
*speech_problems_level (1.62)*
 = {"low"/0.37,"normal"/0.62,"high"/0.0}
*family_implication (2.00)*
 = {"reduce"/0.0,"moderate"/1.0,"high"/0.0}
*children_age (4.50)*
 = {"small"/0.25,"medium"/0.5,"big"/0.0}
We consider five rules for illustrate the inference steps:

- *IF (speech_problems_level is high) and (child_age is medium) and (family_implication is reduce) THEN weekly_session_number is high;*

min (0.00, 0.50, 0.00) = **0.00** for linguistic term **high**

- *IF (speech_problems_level is low) and (child_age is small) and (family_implication is moderate) THEN weekly_session_number is low;*

min (0.37, 0.25, 1.00) = **0.25** for linguistic term **low**

- *IF (speech_problems_level is low) and (child_age is medium) and (family_implication is moderate) THEN weekly_session_number is low;*

min (0.37, 0.50, 1.00) = **0.37** for linguistic term **low**

- *IF (speech_problems_level is normal) and (child_age is small) and (family_implication is moterate)THEN weekly_session_number is normal*

min (0.62, 0.25, 1.00) = **0.25** for linguistic term **normal**

- *IF (speech_problems_level is normal) and (child_age is medium) and (family_implication is moderate)THEN weekly_session_number is normal*

min (0.62, 0.5, 1.00) = **0.50** for linguistic term **normal**

Final confidence coefficients levels are obtained using max function:
- **high** = *max (0.00)* = **0.00**
- **low** = *max (0.25, 0.37)* = **0.37**
- **normal** = *max (0.25, 0.50)* = **0.50**

Each linguistic term of output variable has another representation and in this manner is obtained final graphical representation of weekly_session_number variable. If system user wants to get a single output value, then area center of gravity is calculated. In our case (value 1.62), child must participate at one to two session (but two is preferred).

For knowledge base representation, we use standard Fuzzy Control Language (FCL). In this approach, variables presented above, has following code representation.

```
VAR_INPUT
  speech_problems_level : REAL;
  child_age : REAL;
  family_implication : REAL;
END_VAR

VAR_OUTPUT
  weekly_session_number : REAL;
END_VAR

FUZZIFY speech_problems_level
  TERM low := (0.5, 1) (1, 1) (2,0) ;
  TERM normal := (1,0) (2, 1) (3,0);
  TERM high := (2, 0) (3,1) (3.5,1);
END_FUZZIFY

FUZZIFY family_implication
  TERM reduce := 1;
  TERM moderate := 2;
  TERM high := 3;
END_FUZZIFY

FUZZIFY child_age
  TERM small := (3,1) (5,0);
  TERM medium := (4,0) (5,1) (6,0);
  TERM big := (5.5,0) (7,1) (8,1);
END_FUZZIFY

DEFUZZIFY weekly_session_number
  TERM low := (0.5,1) (1,1) (2,0);
  TERM normal := (1,0) (2,1) (3,0);
  TERM high := (2,0) (4,1);
  ACCU : MAX;
  METHOD : COG;
  DEFAULT := 0;
END_DEFUZZIFY
```

Member functions are represented by specific sets of point (polygonal representation). Following defuzzification methods are available:
- **COG** - Centre Of Gravity;
- **COGS** - Centre Of Gravity for Singletons;
- **COA** - Centre Of Area;
- **LM** - Left Most Maximum;
- **RM** - Right Most Maximum.

OR and AND operators, has different possible formula, specified in FCL file (table I).

TABLE I. THE POSSIBLE FORMULAS FOR LOGIC OPERATORS

| **OR** operator | |
|---|---|
| **keyword** | **algorithm** |
| **MAX** | max $(\mu_1(x), \mu_2(x))$ |
| **ASUM** | $\mu_1(x)+\mu_2(x)-\mu_1(x)*\mu_2(x)$ |
| **BSUM** | min $(1, \mu_1(x)+\mu_2(x))$ |
| **AND** operator | |
| **keyword** | **algorithm** |
| **MIN** | min $(\mu_1(x), \mu_2(x))$ |
| **PROD** | $\mu_1(x)*\mu_2(x)$ |
| **BDIF** | max $(0, \mu_1(x)+\mu_2(x)-1)$ |

## IV. INFERENCE ENGINE AND VALIDATION INTERFACE

In order to implement inference engine, we utilize and adapt a set of Java open-source classes. Used Saznov Java open-source classes are enumerated below:
- class fuzzy.**FuzzyBlockOfRules**
- class fuzzy.**FuzzyEngine**
- class fuzzy.**FuzzyExpression**
- class fuzzy.**FuzzyRule**
- class fuzzy.**Hedge**
  - class fuzzy.**HedgeNot**
  - class fuzzy.**HedgeSomewhat**
  - class fuzzy.**HedgeVery**
- class fuzzy.**LinguisticVariable**
- class fuzzy.**MembershipFunction**
- class java.lang.Exception
  - class fuzzy.**EvaluationException**
  - class fuzzy.**NoRulesFiredException**
  - class fuzzy.**RulesParsingException**

Another complex Java open-source implementation of an inference engine is JFuzzyLogic (figure 9).

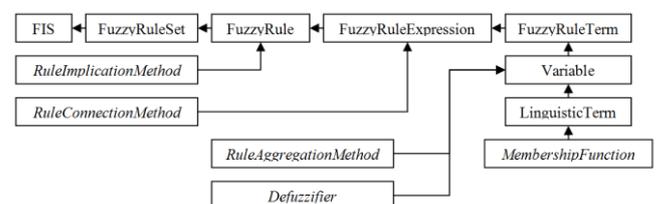

Figure 9 – JFuzzyLogic inference engine architecture

A more detailed view of figure 9 (italic marked class) can be found in table II.

TABLE II. DETALED VIEW OF JFUZZYLOGIC ARCHITECTURE

| |
| --- |
| ***MembershipFunction*** |
| - *MembershipFunctionContinous* |
|     o MembershipFunctionGaussian |
|     o MembershipFunctionGenBell |
|     o MembershipFunctionPiceWiseLinear |
|     o MembershipFunctionSigmoidal |
|     o MembershipFunctionTrapezoidal |
|     o MembershipFunctionTriangular |
| - *MembershipFunctionDiscrete* |
|     o MembershipFunctionGenericSingleton |
|     o MembershipFunctionSingleton |
| ***Defuzzifier*** |
| - *DefuzzifierContinous* |
|     o DefuzzifierCentreOfArea |
|     o DefuzzifierCentreOfGravity |
|     o DefuzzifierLeftMostMax |
|     o DefuzzifierMeanMax |
|     o DefuzzifierRightMostMax |
| - *DefuzzifierDiscrete* |
|     o DefuzzifierCentreOfGravitySingleton |
| ***RuleAggregationMethod*** |
| - RuleAggregationMethodBoundedSum |
| - RuleAggregationMethodMax |
| - RuleAggregationMethodNormedSum |
| - RuleAggregationMethodProbOr |
| - RuleAggregationMethodSum |
| ***RuleImplicationMethod*** |
| - RuleImplicationMethodMin |
| - RuleImplicationMethodProduct |
| ***RuleConnectionMethod*** |
| - RuleConnectionMethodAndBoundedDif |
| - RuleConnectionMethodAndMin |
| - RuleConnectionMethodAndProduct |
| - RuleConnectionMethodOrBoundedSum |
| - RuleConnectionMethodOrMax |
| - RuleConnectionMethodOrProbOr |

In order to validate obtained inference engine, we develop a specific interface (figure 10).

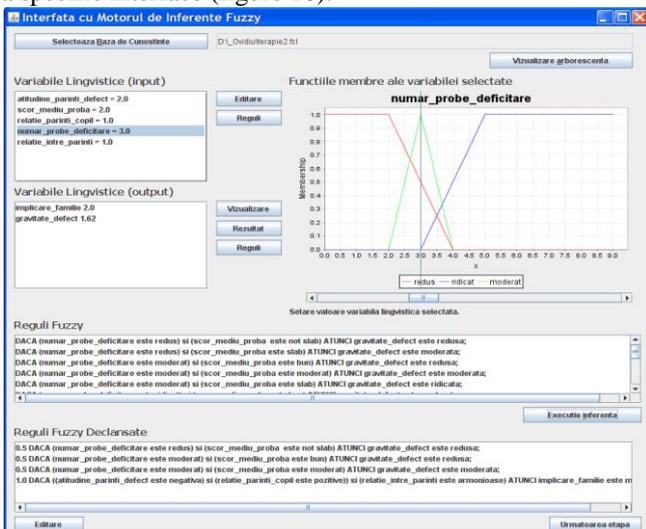

Figure 10 – Expert system validation interface

This interface is detailed presented in papers [3]. We implement over 150 fuzzy rules for control various aspects of personalized therapy (19 variables presented in figure 11). These rules are currently validated by speech therapists and can be modified in a distributed manner.

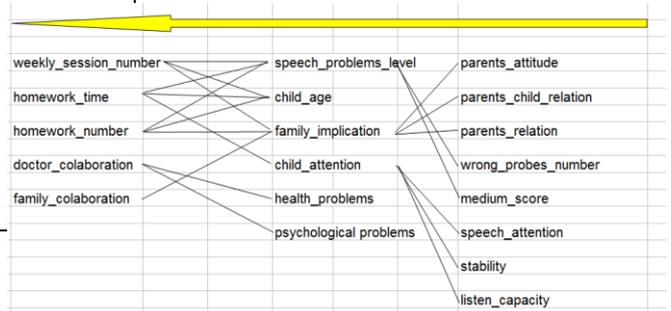

Figure 11 – Fuzzy variables used for expert system

## V. CONCLUSION

In order to help speech therapist activity, we implement a knowledge database with over 150 rules for 19 variables. Fuzzy approach was elected and a validation interface was implemented.

## ACKNOWLEDGMENTS

We must specify that these researches are part of TERAPERS project financed by the National Agency for Scientific Research, Romania, INFOSOC program (contract number: 56-CEEX-II-03/27.07.2006).